\documentclass[letterpaper, 10 pt, conference]{ieeeconf}  

\IEEEoverridecommandlockouts                              

\usepackage{graphicx}
\usepackage{subcaption}
\usepackage{color}
\usepackage{amsmath}
\usepackage{amssymb}
\usepackage{breqn}
\usepackage{verbatim}
\usepackage{balance}

\usepackage[ruled, linesnumbered]{algorithm2e}

\SetCommentSty{mycommfont}

\newcounter{myenum}
	{\end{list}}

\newenvironment{flushitemize}{%
	\begin{list}{$\bullet$}
		{\setlength{\leftmargin}{15pt}}%
		\setlength{\labelwidth}{20pt}
		\setlength{\itemindent}{0pt}
		\setlength{\labelsep}{0.5em}
		\setlength{\itemsep}{1pt}
		\setlength{\parskip}{0pt}
		\setlength{\parsep}{0pt}}
	{\end{list}}

\graphicspath{{imgs/}}

\title{\LARGE \bf
	Tool Macgyvering: Tool Construction Using Geometric Reasoning
}

\author{Lakshmi Nair$^1$, Jonathan Balloch$^1$, and Sonia Chernova$^1$
\thanks{$^1$ Georgia Inst. of Technology, Atlanta, GA. Email: {\tt\small \{lnair3,balloch,chernova\}@gatech.edu}}%
}

\begin{document}
	
	\maketitle
	\thispagestyle{empty}
	\pagestyle{empty}
	
	
	\begin{abstract}
    \textit{MacGyvering} is defined as creating or repairing something in an inventive or improvised way by utilizing objects that are available at hand. In this paper, we explore a subset of Macgyvering problems involving tool construction, i.e., creating tools from parts available in the environment.  We formalize the overall problem domain of tool Macgyvering, introducing three levels of complexity for tool construction and substitution problems, and presenting a novel computational framework aimed at solving one level of the tool Macgyvering problem, specifically contributing a novel algorithm for tool construction based on geometric reasoning.  We validate our approach by constructing three tools using a 7-DOF robot arm.
    \end{abstract}
	
	
	\section{Introduction}
	
	Intelligence is often best expressed through creative problem solving - a skill that humans frequently depend on, especially in high-stress or time constrained scenarios. In the Apollo 13 incident of 1970, a carbon dioxide filter creatively constructed out of a sock, a plastic bag, book covers, and duct tape helped save the lives of the three astronauts on board \cite{cass2005apollo}. Solving problems by constructing new tools from available objects is colloquially referred to as ``Macgyvering''. The term originated from the popular TV series \textit{Macgyver}, which featured a secret service agent who used common objects available to him in the absence of required tools in order to escape difficult situations. Unlike humans, today's robots are limited to using predefined available tools, although some prior work has explored \textit{tool substitution} as a means of problem solving and adaptation \cite{abelha2016model,schoeler2016bootstrapping,boteanu2015towards}. Our research focuses on \textit{tool construction}, contributing a computational framework that enables a robot to construct, or Macgyver, tools out of parts available in the environment.  To the best of our knowledge, this is the first work to demonstrate tool construction on a physical robot.
	
\begin{figure}[t]
		\centering
		\includegraphics[width=0.48\textwidth]{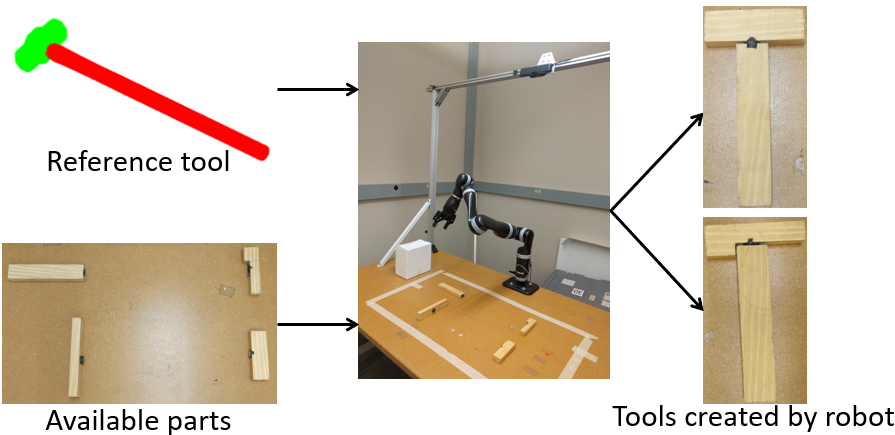}
		\captionsetup{width=\linewidth}
		\caption{Tool creation: Given a reference tool, the robot constructs a substitute tool out of available parts.}
		\label{fig:Overview}
	\end{figure}

	
	Tools are defined as objects that extend the physical influence of the agent \cite{jones1973tool}, and the construction of tools to solve problems is a particularly interesting challenge for intelligent robotic systems because it demonstrates a more sophisticated level of intelligence than simple tool use. Existing work in psychology has shown that tool-making emerges notably later than tool use in children, and that physical tool making is preceded by the step of imagining some ``canonical tool" suitable for the task \cite{beck2011making}. In our work, we refer to this conceived tool as the \textit{reference tool}. Given a 3D model of a reference tool, our system reasons about the geometric properties (e.g., shapes, sizes, attachment points) of available parts to construct a substitute tool (Fig. \ref{fig:Overview}). Our work makes two contributions: 1) we formalize the overall problem domain of tool Macgyvering, introducing 3 levels of complexity for tool construction and substitution problems, and 2) we introduce a computational framework aimed at solving one level of the tool Macgyvering problem, specifically contributing a novel algorithm for tool construction based on geometric reasoning.  We validate our approach by constructing three tools using a 7-DOF robot arm.
	
	\section{Related Work}
	Tool-making and tool use have been widely studied in animals.  Mammals, such as chimpanzees \cite{boesch1990tool} and bonobos \cite{toth1993pan}, and birds, such as rooks \cite{bird2009insightful} and blue jays \cite{jones1973tool}, have been shown to create and use simple tools to retrieve food. In \cite{beck2011making}, the authors identify two key aspects of tool-making: \textit{tool innovation} and \textit{tool manufacturing}.  Tool innovation is the process of imagining the ideal tool required for a given task, and tool manufacturing is the process of physical transformation of materials/parts into a tool.
	
	In the context of robotics, prior work in tool substitution \cite{abelha2016model, schoeler2016bootstrapping} has used visual reasoning for finding tool substitutes. Further, recent work has explored Macgyvering and the use of environmental objects for problem solving. Sarathy and Scheutz \cite{sarathy2017macgyver} proposed a theoretical formulation of Macgyvering; their work differs from ours in that it does not reason about visual/physical properties of objects in tool creation. In \cite{erdogan2013planning}, the authors introduce techniques for reasoning about construction of functional structures for navigation. Further work explored use of environmental objects as simple machines \cite{stilman2014robots, levihn2014using}. In \cite{wicaksono17towards}, the authors focus on creation of novel tools using 3D printing.  More recently, Choi et al. \cite{choi2018creating} extended the cognitive architecture ICARUS to support the creation and use of tools in abstract planning scenarios. Our work differs from these approaches in that we consider the geometric properties of tools, and demonstrate robot tool construction from diverse environmental objects, including physical validation of the construction.
	
	\begin{table*}[t!]
		\centering
		\includegraphics[width=0.9\textwidth]{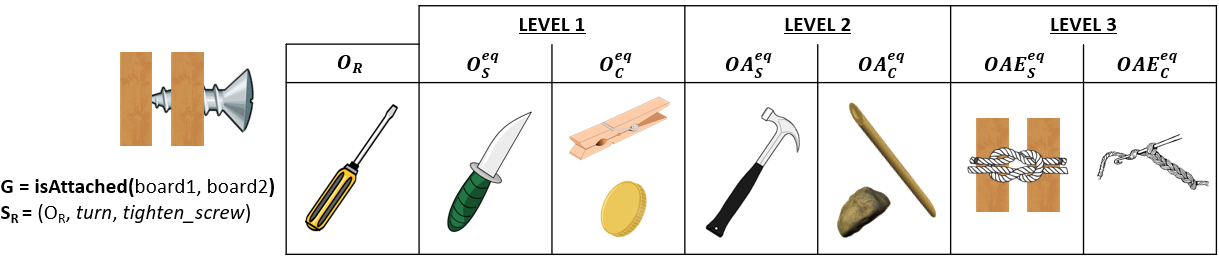}
		\captionsetup{width=\linewidth}
		\caption{Levels of Macgyvering for the task of attaching two wooden boards. The reference solution ($S_R$) is to \textit{turn} a screw with a \textit{screwdriver} to \textit{tighten} it. The affordance equivalences ($O^{eq}$, $OA^{eq}$ and $OAE^{eq}$) are indicated for each level for both tool substitution (S) and construction (C).} 
		\label{fig:MG_levels_objects}
	\end{table*}
	
	\begin{figure}[t]
		\centering
		\includegraphics[width=0.35\textwidth]{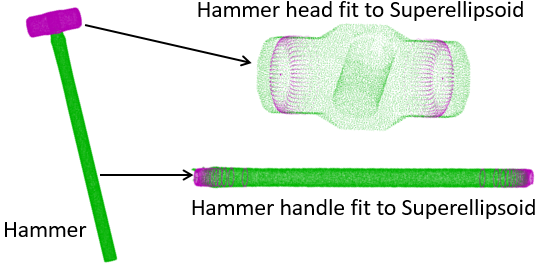}
		\captionsetup{width=\linewidth}
		\caption{Superquadrics fitted to a hammer: the hammer head and handle parts (right) are both represented by Superellipsoids (purple) obtained by fitting SQs to the tool segments.}
		\label{fig:SQ_examples}
	\end{figure}
	
	One of the key design choices within our framework is the underlying representation used for tools.  We adopt the tool representation proposed by Abelha et al. \cite{abelha2016model}, who demonstrated the superior performance of Superquadrics (SQ) over other techniques in the context of object substitution.  Superquadrics refer to the family of geometric shapes that includes quadrics, but allows for arbitrary powers instead of just power of two \cite{barr1981superquadrics}. Fig. \ref{fig:SQ_examples} shows an example of a SQ model of a hammer, where both the hammer head and handle are represented by a class of SQs called Superellipsoids. In our work, we show that the SQ representation of tools allows the robot to select and attach geometrically appropriate pieces when constructing new tools. 
	
	\section{Levels of Tool Macgyvering}
	
	The problems of tool use, tool substitution and tool construction are all related to the representation of tool affordances.  Below, we first define and characterize tool affordances, and then show that complexity levels of the tool Macgyvering problem can be formalized based on the extent to which the tool affordance representation must be adapted.
	
	\subsection{Macgyvering as a function of object affordances}
	
	Problems that are solved using tool construction and substitution are defined by their two primary constituents: a \textit{task goal} $G$ and a \textit{solution affordance} $S$. We define goals as persistent relationships between objects, such as ``\textit{isAttached}(painting, wall)'', and affordances as action possibilities available to the agent for a given object \cite{gibson1977perceiving}. Computationally, affordances are defined as unique relationships between objects $(O)$, the actions those objects can be used for $(A)$, and the effects $(E)$ of applying the actions with the objects \cite{lopes2007affordance}. We use the tuple $S = (O, A, E)$ to denote an affordance solution for goal $G$. 
	
	Prior work on affordance representations \cite{andries2018affordance}, has defined \textit{object equivalence} ($O^{eq}$) as occurring when the same action applied to two different objects generates an equivalent effect.  Similarly, \textit{object-action equivalence} ($OA^{eq}$) is defined as occurring when different actions applied to different objects result in equivalent effects.   We introduce a third equivalence class, \textit{object-action-effect equivalence} ($OAE^{eq}$), as occurring when different actions applied to different objects generate different effects, but that those effects accomplish the same task goal $G$. Table \ref{fig:MG_levels_objects} presents an example of all three levels of equivalence for the task goal \textit{G=isAttached}(board1, board2) and reference solution $S_R=(screwdriver,turn,tighten\_screw)$. $OAE^{eq}$ occurs when the goal of attaching the boards is achieved, but without using the original tool, action or effect, such as by tying the boards together with a rope instead of tightening the screw. Below, we show that tool Macgyvering problems can be broken down into three classes, based on which elements of the reference solution $S_R$ are assumed to remain unchanged.
	
	\subsection{Macgyvering Levels}
	We define three levels of Macgyvering based on the above equivalence classes, and show how the relative complexity of tool-based problems can be assessed directly from the affordance equivalency representation.  Each Macgyvering level consists of two variants based on whether reference objects are substituted or constructed. \textit{Tool substitution}, denoted by the subscript $_S$, refers to the case in which an existing tool can be used to replace a reference tool. \textit{Tool construction} on the other hand, denoted by the subscript $_C$, refers to the case in which no substitute for the reference tool exists, and a new tool must be constructed.  The definitions below, also summarized in Table \ref{fig:MG_levels}, are in reference to a task with goal $G$ and a reference solution defined by $S_R=(O_{R}, A_{R}, E_{R})$.
	
	\begin{table}[t]
		\centering
		\includegraphics[width=0.45\textwidth]{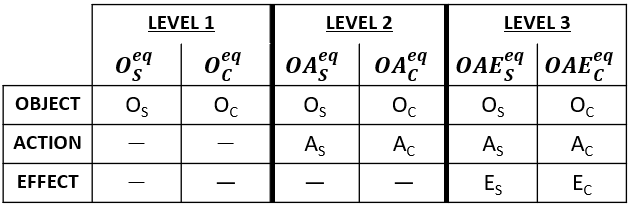}
		\captionsetup{width=\linewidth}
		\caption{The 3 Levels of Macgyvering; ``---'' indicates the action or effect of the reference tool is being preserved.}
		\label{fig:MG_levels}
	\end{table}

	\smallskip
	\textbf{Level 1:} \textbf{\textit{object equivalence solutions}}.  The first level addresses tasks with solutions that can be achieved either through object substitution ($O_S^{eq}$) or object construction ($O_C^{eq}$) alone, where the newly found or created object shares the action and effect of the reference solution. Thus, the solution affordance $S$ for Level 1 is $(O_{S}, A_{R}, E_{R})$ or $(O_{C}, A_{R}, E_{R})$. As shown in Table \ref{fig:MG_levels_objects}, in the case of $O_S^{eq}$ a knife can be used in place of a screwdriver to \textit{turn} and tighten a screw, and in the case of $O_C^{eq}$, a clothespin and a coin can be combined to create a tool identical in function to the screwdriver.
	
	\smallskip
	\textbf{Level 2:} \textbf{\textit{object-action equivalence solutions}}. The second level addresses tasks for which no viable object equivalence solution exists, and the goal is achieved by substituting, $OA_S^{eq}$, or constructing, $OA_C^{eq}$, a tool with a non-reference action where the object-action pair yields the same effect as the reference solution. Thus, for Level 2 $S = (O_{S}, A_{S}, E_{R})$ or $S = (O_{C}, A_{C}, E_{R})$, where $A_{S}, A_{C} \neq A_{R}$.  In our example of attaching two boards, in the case of $OA_S^{eq}$, a hammer can be used to push the screw in with a \textit{hitting} action instead of using the screwdriver with a \textit{turn} action. Both actions accomplish the effect of tightening the screw. In the case of $OA_C^{eq}$, the same hitting action would be used, but the hammer would need to be constructed, such as by combining a stick and a rock.
	
	\smallskip
	\textbf{Level 3:} \textbf{\textit{object-action-effect equivalence solutions}}. The third level addresses tasks for which no viable object-action equivalence solution exists, and the goal $G$ is achieved by substituting, $OAE_S^{eq}$, or constructing, $OAE_C^{eq}$, a tool with a non-reference action and effect. For Level 3, $S = (O_{S}, A_{S}, E_{S})$ or $S = (O_{C}, A_{C}, E_{C})$, where $A_{S}, A_{C} \neq A_{R}$ and $E_{S}, E_{C} \neq E_{R}$. In our running example, in the case of $OAE_S^{eq}$ a rope can be used to \textit{tie} two pieces of wood together, and in the case of $OAE_C^{eq}$ a rope would first be knitted out of strands of yarn and then used to \textit{tie}. Both scenarios no longer accomplish the effect of tightening the screw, but accomplish the task goal of attaching the boards.
	
	\smallskip
	
	As can be seen from the examples, developing a robotic system that can solve the Macgyvering problem becomes progressively harder at each level as the reasoning requires a deeper understanding of the domain, desired task goal, the actions afforded by each object, and their resulting effects. 
	
	\begin{figure*}[t]
		\centering
		\includegraphics[width=0.99\textwidth]{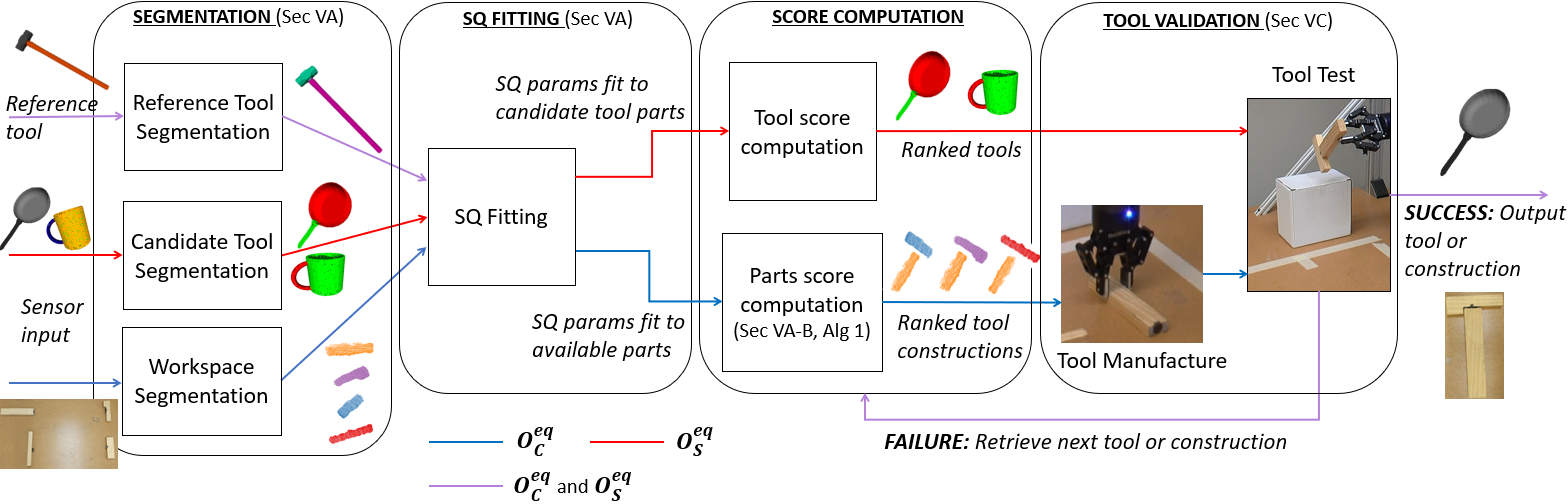}
		\captionsetup{width=\linewidth}
		\caption{Framework for Level 1 Macgyvering, shown with the example of a hammer. This figure aims to highlight the integration of tool substitution and construction within a single framework. The $O^{eq}_S$ pipeline follows from \cite{abelha2016model}. Our novel contribution and focus in this work is the tool construction (blue and purple) pipeline.} 
		\label{fig:MG_module}
	\end{figure*}
	
	\section{Framework for Level 1 Macgyvering} 
	
	In this section, we present a computational framework for Level 1 Macgyvering. Fig. \ref{fig:MG_module} presents an overview, with red lines denoting the tool substitution $O_S^{eq}$ pipeline, blue lines denoting the tool construction $O_C^{eq}$ pipeline, and purple denoting shared components.  
	
	\textit{Tool Substitution:} The object substitution scenario follows the pipeline described by Abelha et al. \cite{abelha2016model}, wherein the system is provided with a reference tool and a set of candidate tools. The reference and candidate tools are segmented into their constituent parts (hereafter, ``tool components''), and the resulting components are fitted with SQs. Candidate tools are then scored and ranked based on the similarity of their components to the reference tool components. We further extend this work by incorporating a tool validation phase in which the robot physically tests that the substitution tool performs the necessary function; failure of this test causes lower ranked substitution candidates to be considered.  
	
	
	
	\textit{Tool Construction:} In the tool construction scenario, the system is provided with a reference tool and a set of parts available for construction (hereafter, ``candidate parts'').  As before, the reference tool is segmented, then SQs are fitted to each reference tool component and candidate part. For the tool construction phase of the pipeline, we introduce a metric that takes into account the shape, size and possible configurations of the candidate parts and outputs a ranked list of possible tool constructions. The robot then uses the ranked list to construct candidate tools, which it validates on the target task.    
	
	\smallskip
	
	

Although only the tool construction pipeline serves as the main contribution of this paper, we present it alongside the tool substitution elements to highlight the synergy of our work with prior methods in the field.  In the following section, we present the details of the tool construction methodology.

	\section{Tool Construction $O_C^{eq}$}
	
	We formulate the tool construction problem as follows:
	
	\smallskip
	\textit{``Given a reference tool $R$, and a set $C$ of $n$ candidate parts, how can the robot reason about the geometric properties and attachment points of the available parts in order to construct a replacement tool that can be used to accomplish task goal $G$ using the same action and effect as the reference solution?''}
	\smallskip 
	
	We define \textit{attachment points} as locations at which objects can be attached together.  In general, attachments can result in fixed, revolute or prismatic connections \cite{shiraki2014modeling} achieved through insertion, glueing, duct tape, etc. In this work, we apply our framework in two scenarios: i) when a predefined library exists that specifies the attachment points available for each candidate part, as in \cite{bessler2018owl}, and ii) when no attachment information is known. 
	
	Given $n$ candidate construction parts, the total space of configurations or permutations for the $n$ parts is $^nP_2 + ... + ^nP_n$, assuming at least two and at most $n$ parts must be combined to construct the tool, resulting in a prohibitively large problem space that is combinatorial in the number of candidate parts. In this work, we make the simplifying assumption that the number of parts required to construct the tool is equal to the number of reference tool components\footnote{Most tools have two key components, grasp part and action part \cite{myers2015affordance, abelha2017learning}}, thus, given $m$ tool components, we obtain a state space of $^nP_m$. Below, we detail our tool construction approach.

	\begin{algorithm}[tb]
		
		\SetKwInOut{Input}{input}\SetKwInOut{Output}{output}
		
		\Input{importance weights $\lambda_1, \lambda_2, \lambda_3, \lambda_4$\\SQ params,$T=permute(C,m)$}
		\Output{$T^*$, $Att$}
		\BlankLine
		
		$E = [], Att = []$
		
		\For{$i\gets1$ \KwTo $|T|$}{
			
			\For{$j\gets1$ \KwTo $m$}{
				$e_{shape}^{T_i} \stackrel{+}{=} |shape(r_j) - shape(T_{ij})|$ 
				
				$e_{scale}^{T_i} \stackrel{+}{=} |scale(r_j) - scale(T_{ij})|$ 
				
				\For{$k\gets1$ \KwTo $m$}{
					
					\uIf{$k \neq j$}{
						$e_{ratio}^{T_i} \stackrel{+}{=} |rel(r_j,r_k) - rel(T_{ij},T_{ik})|$
					}
				}
			}
			$e_{att}^{T_i}, A^{T_i}_{closest} = AttachmentFit(T_i)$ \tcp{Alg 2}
			
			$e_{const}^{T_i} = \lambda_1 e_{scale}^{T_i} + \lambda_2 e_{shape}^{T_i} + \lambda_3 e_{ratio}^{T_i} + \lambda_4 e_{att}^{T_i}$
			
			$E.append(e^{T_i})$
			
			$Att.append(A^{T_i}_{closest})$
			
		}
		
		$T^* = sort(T, E)$ \tcp{Sort $T$ based on $E$}
		\Return $T^*, Att$
		
		\caption{Parts Score Computation}
	\end{algorithm}
	
	\subsection{Segmentation, SQ Fitting and Part Score Computation}
	
	The tool construction pipeline begins with segmentation, which enables the system to identify the basic components of the reference tool and the candidate parts in the robot's workspace.  Our reference tool model is obtained from the ToolWeb dataset \cite{abelha2017learning}, and we use Triangulated Surface Mesh Segmentation\footnote{The implementation was provided by the CGAL library} to segment the tool (e.g., into handle and head of a hammer). The resultant reference tool components are denoted as an ordered tuple $R = (r_1, r_2, ..., r_m)$. We use plane subtraction and Sample Consensus Segmentation (SAC)\footnote{The implementation was provided by the PCL library} to identify the candidate parts available to the robot using RGB-D data from a camera mounted over the table. We denote the resulting candidate parts as $C = \{c_1, c_2, ..., c_n\}$. 
	
	Given $R$ and $C$, we next fit SQ\footnote{the presented framework can also be extended to other representations, e.g., ESF and SHOT \cite{schoeler2016bootstrapping}} models to each reference tool component and candidate part.  Each SQ model has 13 parameters: 3 for scale in each dimension, 2 for shape variance, 3 for Euler angles, 2 for tapering parameters and 3 for the central point/mean. We use Levenberg-Marquardt optimization to find the best fit parameters for each object \cite{abelha2017learning}.  We then use Algorithm 1 for computing the part scores.  As input, the algorithm is provided a list of importance weights $\lambda_{1-4}$, SQ parameters of parts and a list of tuples $T$ consisting of all possible candidate configurations generated through the permutation $^nP_m$ of candidate parts in $C$; ordering of parts within each tuple, determines pairwise matching between $c_i$ and $r_i$. The output of Algorithm 1 is a list of candidate builds $T^*$ sorted by a computed vector of error values $e_{const} \in E$, and a list of attachment points $Att$ to use for combining the parts. To compute $e_{const} \in E$, we consider the following four metrics:
	
	\begin{flushitemize}
		
		\item per-component \textbf{\textit{shape}} fit, based on absolute difference between the two shape SQ parameters (denoted by 2D vector $shape$) of candidate parts and corresponding reference components (line 4).
		
		\item per-component \textbf{\textit{size}} fit, based on of absolute difference between the 3 scaling parameters (denoted by 3D vector $scale$) of candidate parts and their corresponding reference tool components scaled to real-world size (line 5). 
		
		\item pairwise component \textit{\textbf{proportionality}} fit, calculated based on the relative scale ratios (denoted by $rel$) between reference tool components and candidate parts (line 8)  Example: hammer handle may be $3x$ longer than the head.
		
		\item if attachment point information is known, an attachment score encoding the \textit{\textbf{proximity of available attachment points}} to the target attachment locations (line 11, details in Section \ref{sec:attachments}) is computed for the different possible part configurations of $T_i$. 
	\end{flushitemize}
	
	Each of the above metrics produces an error value indicating the magnitude of the deviation of the candidate part combinations from the reference tool.  We then compute an aggregate error term $e_{const}$ as a weighted sum (line 12) of all four error terms using $\lambda$ importance weights\footnote{$\lambda$ terms were manually chosen in this work}. The list of candidate part configurations are then sorted by their associated error value $e_{const}$, from lowest to highest.


	\subsection{Attachments With and Without Known Points}
	\label{sec:attachments}
	
	The presented framework is able to construct tools both when a list of predefined attachment points is available, and when it is not (Algorithm 2).  In both cases, the process begins by aligning the components of the candidate tool $T_i$ in a configuration consistent with $R$ (line 2); we use Principal Component Analysis (PCA) to orient part point clouds w.r.t. the reference tool, resulting in a set of alignments $T_i'$. We approximate the intersections of the point clouds in each alignment by calculating the centroid of closest points between the point clouds (line 3). The resultant set of centroids, $P$, is the candidate list of attachments we want to make.
	
	
	
\begin{figure}[t]
		\centering
		\includegraphics[width=0.46\textwidth]{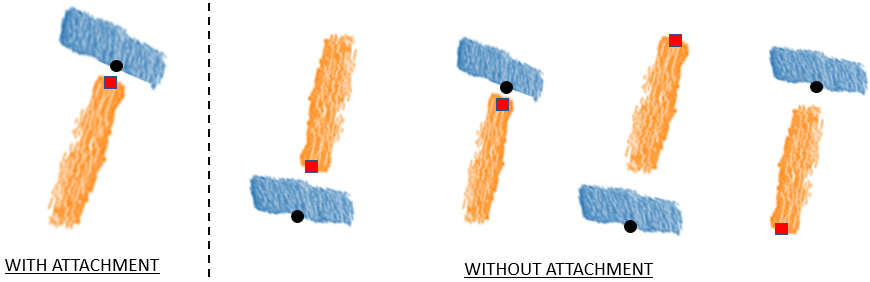}
		\captionsetup{width=\linewidth}
		\caption{The squares and circles indicate attachment points. With unknown attachments, the robot uses $P$ to explore different configurations. With known attachments, the robot directly attempts the correct configuration (shown to the left).}
		\label{fig:attach}
	\end{figure}

	If no part attachment information is known, the set $P$ is used to guide the robot in constructing the tool.  Specifically, the robot will attempt to attach parts at the locations specified by $P$, and then verify whether attachment was successful before proceeding (see Fig \ref{fig:attach} and accompanying video\footnote{https://www.youtube.com/channel/UCxnm8iu1TS75YNXcAiI-nEw}).  This search process enables the robot to explore possible attachments until a successful part combination is found, however, leading to more tool construction attempts.  Alternatively, if attachment information is provided, it is incorporated into $T_i$ ranking by computing an error score based on the Euclidean distance between points in $P$ and the closest attachment locations on each part $c_j$, in each alignment $t_i \in T_i'$ (lines 5-9). The resulting error value, $e_{att}$, is combined with the other error metrics, described above, to rank candidate parts and configurations. If a part in $t_i$ is known to have no attachment points, $e^{T_i}_{att} = \infty$.
	
	
	\begin{algorithm}[tb]
		\SetKwInOut{Input}{input}\SetKwInOut{Output}{output}
		
		\Input{candidate tool parts $T_i$, attachments $A$}
		\Output{$e_{att}^{T_i}$, $A_{closest}$}
		\BlankLine
				
		$e^{T_i}_{att}=0,$
		$A_{closest} = []$
		
		\tcp{pose and orient parts in $T_i$ in ref. to $R$}   
		$T_i' = Align(T_i, R)$ 
	          
		$P = ComputeIntersections(T_i')$ 
		
		\uIf{$A \neq \varnothing$}{
			\ForEach{$t_i \in T_i', c_j \in t_i$}{
				$a = ClosestAttachments(P, c_j, A)$
						
				$e^{T_i}_{att} += \|P, a\|$
						
				$A_{closest}.append(a)$
			}
		}
		\Else{
			\Return $e^{T_i}_{att}, P$ \tcp{Attachments unknown}
		}
		
		\Return $e^{T_i}_{att}, A_{closest}$
		
		\caption{Attachment Fit}
	\end{algorithm}

	
	
	\begin{figure}[t]
		\centering
		\includegraphics[width=\columnwidth]{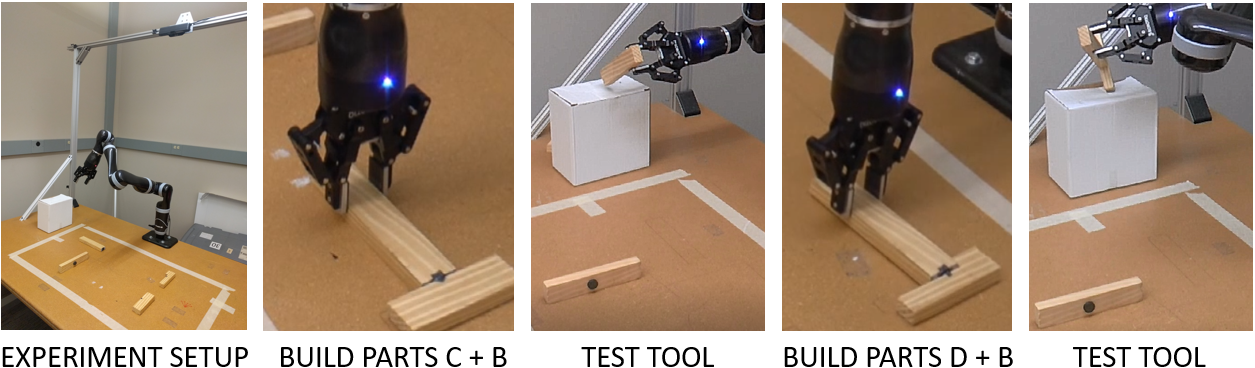}
		\caption{Hammer construction, first using parts C and B (failure due to tool breaking apart on impact), and second using parts D and B (success).}
		\label{fig:expt}
	\end{figure}
	
	\begin{table}[t]
		\centering
		\includegraphics[width=0.486\textwidth]{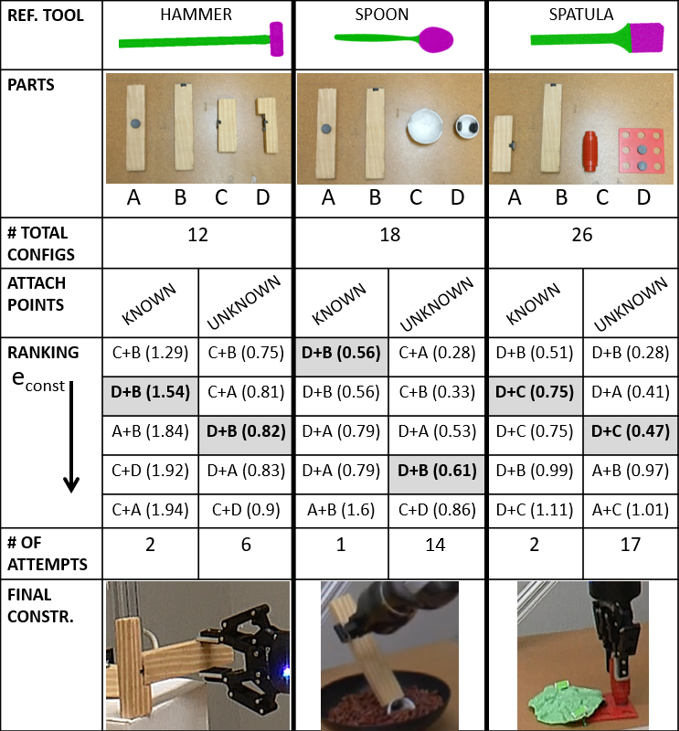}
		\captionsetup{width=\linewidth}
		\caption{For the 3 reference tools, we show the candidate parts and total possible configurations. ``Ranking'' shows part combinations ranked from best to worst. The bold and shaded scores correspond to the final working tool. Number of physical attempts are also shown for each case.}
		\label{fig:examples_table}
	\end{table}
	
	
	
	\subsection{Tool Validation with Manufacturing and Testing} 
	Given the ranked list of potential tool constructions obtained above, the robot next constructs a tool by joining the components specified by the best-rated configuration using the attachment points (output of Algorithm 1).  The robot then evaluates the constructed tool for its task suitability by applying the desired action on the tool. In this work, we assume that the robot is given the required action for each object, and that it can observe whether the tool succeeded.  Alternatively, this information could be learned from demonstrations \cite{rana2017towards}, including, if necessary, adapting the original action to fit the dimensions of the new tool \cite{fitzgerald2014representing,gajewski2018adapting}. 
	

	\section{Experimental Validation}
	\label{sec:results}
To validate our tool construction approach, we constructed three tools: a hammer, spoon and spatula. Each tool consisted of two components ($m = 2$), and the robot was given $n = 4$ candidate parts. Reference tool models were acquired from the ToolWeb dataset \cite{abelha2017learning}. The manually-set weight parameters $\Lambda = \{1, 1, 5, 5\}$ ($e_{shape}, e_{scale}, e_{ratio}, e_{att}$) worked well across all three tasks. We use magnets as attachments for the parts, and seek to validate the performance of our approach both when attachment points are known and unknown. 




Fig. \ref{fig:expt} shows the robot's workspace and a typical build-and-test cycle of tool building, demonstrated on the hammer. All three tasks, and their results, are summarized in Table \ref{fig:examples_table}, including: the reference tool used, list of available candidate parts, the number of possible configurations of parts given the attachment points, the part combinations that were attempted and that succeeded (shaded, bold) for each build, along with their corresponding $e_{const}$ value, the number of attempts the robot required to build the tool, and the final constructed tool.  Each tool design has a single working configuration, but tests different aspects of our approach. We first describe our results for the case of known attachments, and then unknown attachments. The avg. computation time for $e_{const}$ was only 15.44 seconds and we note other key insights in each case.


\subsection{Tool Construction with Known Attachments}

\textit{Hammer:} Candidate parts for the hammer consisted of four wooden blocks of various dimensions, each with a single attachment point. The highest ranked candidate tool (lowest error) consisted of parts B and C, which most closely resemble a hammer shape.  However, the resulting magnet connection was not strong enough to withstand the impact of the robot hitting a box during the tool validation phase and the tool broke.  Subsequently, the robot attempted to combine the next ranked set of parts, B and D. The resulting tool passed testing because the curved shape of part D, while less closely resembling a real hammer, provided needed support during impact. This demonstrates the robot's ability to validate the tool beyond only geometric fit.  

\textit{Spoon:} Candidate parts for the spoon consisted of two wooden handles and two scoops; one scoop had two attachment points and the other had zero.  Using available attachment information, the robot correctly ignored the scoop with no magnets (which was a closer match by shape alone), and correctly chose to combine parts B and D to construct a spoon that successfully scooped beans out of a bowl. This demonstrates the robot's ability to reason about 1) symmetric constructions (parts D+B ranked first and second due to symmetric attachments) and 2) absence of attachments.  

\textit{Spatula:} Candidate parts for the spatula consisted of three possible handles and one flat piece with two attachments.  The robot first attempted to combine parts B and D, however, a successful connection failed due to polar orientation of the magnets.  The robot recovered from this problem by successfully combining parts C and D, even though the resulting shape was less similar to a spatula than the original attempt.  The resulting tool was successfully used to pick up a piece of plastic lettuce. This demonstrates the robot's capability to 1) create tools that vastly deviate in terms of geometry but still accomplish the function, and 2) reason about multiple attachment configurations (part combinations D+C repeat for the different attachment configurations).  
	
	
\subsection{Tool Construction with Unknown Attachments}
The results in Table \ref{fig:examples_table} highlight several key differences in the construction process with and without known attachments.  When the robot does not have knowledge of attachment points a priori, the construction process is guided by the candidate attachments in $P$ (Section \ref{sec:attachments}). As a result, the robot attempts multiple attachments per configuration.  For example, the solution for the spoon was found using the fourth part combination (D+B) after 14 construction attempts.  Thus, identifying a valid combination of parts using shape information alone is significantly more challenging than when attachment points are known.  However, in all cases the robot successfully identifies a solution within the first 4 configurations (out of possible 12, 18 or 26), validating that the $e_{const}$ ranking is successful at guiding the search.

	\section{Conclusions and Future Work} 
	\label{sec:conclusion}
	
In this work, we have contributed a novel computational framework for tool construction based on geometric reasoning, which we have demonstrated on the construction of three tools, both when part attachment information is known and unknown.  In all cases, the robot efficiently constructed the tool, exploring only a small percentage of all possible part combinations.  Also, we have introduced a formalization of tool Macgyvering, to guide future research in this area by defining varying levels of difficulty for this problem.   Future work should address the following limitations of the current approach.  Two key assumptions, the current limitation that the number of candidate parts must equal the number of tool components, and that $\Lambda$ parameters are hand-coded, should be relaxed. Further, as tool complexity is increased, metrics and heuristics should be explored that maintain the computational tractability of the search problem. Finally, the tool representation should be extended to include a broader range of properties, including material and density.

	
	
	
	
	\section*{Acknowledgments}
	This work is supported in part by NSF IIS 1564080 and ONR N000141612835.
	
	\bibliographystyle{./IEEEtran}
	\balance
	\bibliography{references}  

\begin{thebibliography}{10}
\providecommand{\url}[1]{#1}
\csname url@rmstyle\endcsname
\providecommand{\newblock}{\relax}
\providecommand{\bibinfo}[2]{#2}
\providecommand\BIBentrySTDinterwordspacing{\spaceskip=0pt\relax}
\providecommand\BIBentryALTinterwordstretchfactor{4}
\providecommand\BIBentryALTinterwordspacing{\spaceskip=\fontdimen2\font plus
\BIBentryALTinterwordstretchfactor\fontdimen3\font minus
  \fontdimen4\font\relax}
\providecommand\BIBforeignlanguage[2]{{%
\expandafter\ifx\csname l@#1\endcsname\relax
\typeout{** WARNING: IEEEtran.bst: No hyphenation pattern has been}%
\typeout{** loaded for the language `#1'. Using the pattern for}%
\typeout{** the default language instead.}%
\else
\language=\csname l@#1\endcsname
\fi
#2}}

\bibitem{cass2005apollo}
S.~Cass, ``Apollo 13, we have a solution,'' \emph{IEEE Spectrum On-line, 04},
  vol.~1, 2005.

\bibitem{abelha2016model}
P.~Abelha, F.~Guerin, and M.~Schoeler, ``A model-based approach to finding
  substitute tools in 3d vision data,'' in \emph{Robotics and Automation
  (ICRA), 2016 IEEE International Conference on}.\hskip 1em plus 0.5em minus
  0.4em\relax IEEE, 2016, pp. 2471--2478.

\bibitem{schoeler2016bootstrapping}
M.~Schoeler and F.~W{\"o}rg{\"o}tter, ``Bootstrapping the semantics of tools:
  Affordance analysis of real world objects on a per-part basis,'' \emph{IEEE
  Transactions on Cognitive and Developmental Systems}, vol.~8, no.~2, pp.
  84--98, 2016.

\bibitem{boteanu2015towards}
A.~Boteanu, D.~Kent, A.~Mohseni-Kabir, C.~Rich, and S.~Chernova, ``Towards
  robot adaptability in new situations,'' in \emph{2015 AAAI Fall Symposium
  Series, Arlington, VA: AAAI Press}, 2015.

\bibitem{jones1973tool}
T.~B. Jones and A.~C. Kamil, ``Tool-making and tool-using in the northern blue
  jay,'' \emph{Science}, vol. 180, no. 4090, pp. 1076--1078, 1973.

\bibitem{beck2011making}
S.~R. Beck, I.~A. Apperly, J.~Chappell, C.~Guthrie, and N.~Cutting, ``Making
  tools isn’t child’s play,'' \emph{Cognition}, vol. 119, no.~2, pp.
  301--306, 2011.

\bibitem{boesch1990tool}
C.~Boesch and H.~Boesch, ``Tool use and tool making in wild chimpanzees,''
  \emph{Folia primatologica}, vol.~54, no. 1-2, pp. 86--99, 1990.

\bibitem{toth1993pan}
N.~Toth, K.~D. Schick, E.~S. Savage-Rumbaugh, R.~A. Sevcik, and D.~M. Rumbaugh,
  ``Pan the tool-maker: investigations into the stone tool-making and
  tool-using capabilities of a bonobo (pan paniscus),'' \emph{Journal of
  Archaeological Science}, vol.~20, no.~1, pp. 81--91, 1993.

\bibitem{bird2009insightful}
C.~D. Bird and N.~J. Emery, ``Insightful problem solving and creative tool
  modification by captive nontool-using rooks,'' \emph{Proceedings of the
  National Academy of Sciences}, vol. 106, no.~25, pp. 10\,370--10\,375, 2009.

\bibitem{sarathy2017macgyver}
V.~Sarathy and M.~Scheutz, ``The macgyver test-a framework for evaluating
  machine resourcefulness and creative problem solving,'' \emph{arXiv preprint
  arXiv:1704.08350}, 2017.

\bibitem{erdogan2013planning}
C.~Erdogan and M.~Stilman, ``Planning in constraint space: Automated design of
  functional structures,'' in \emph{Robotics and Automation (ICRA), 2013 IEEE
  International Conference on}.\hskip 1em plus 0.5em minus 0.4em\relax IEEE,
  2013, pp. 1807--1812.

\bibitem{stilman2014robots}
M.~Stilman, M.~Zafar, C.~Erdogan, P.~Hou, S.~Reynolds-Haertle, and G.~Tracy,
  ``Robots using environment objects as tools the ‘macgyver’ paradigm for
  mobile manipulation,'' in \emph{Robotics and Automation (ICRA), 2014 IEEE
  International Conference on}.\hskip 1em plus 0.5em minus 0.4em\relax IEEE,
  2014, pp. 2568--2568.

\bibitem{levihn2014using}
M.~Levihn and M.~Stilman, ``Using environment objects as tools: Unconventional
  door opening,'' in \emph{Intelligent Robots and Systems (IROS 2014), 2014
  IEEE/RSJ International Conference on}.\hskip 1em plus 0.5em minus 0.4em\relax
  IEEE, 2014, pp. 2502--2508.

\bibitem{wicaksono17towards}
H.~Wicaksono and C.~S.~R. Sheh, ``Towards explainable tool creation by a
  robot,'' in \emph{IJCAI-17 Workshop on Explainable AI (XAI)}, p.~63.

\bibitem{choi2018creating}
D.~Choi, P.~Langley, and S.~T. To, ``Creating and using tools in a hybrid
  cognitive architecture,'' 2018.

\bibitem{barr1981superquadrics}
A.~H. Barr, ``Superquadrics and angle-preserving transformations,'' \emph{IEEE
  Computer graphics and Applications}, vol.~1, no.~1, pp. 11--23, 1981.

\bibitem{gibson1977perceiving}
J.~J. Gibson, ``Perceiving, acting, and knowing: Toward an ecological
  psychology,'' \emph{The Theory of Affordances}, pp. 67--82, 1977.

\bibitem{lopes2007affordance}
M.~Lopes, F.~S. Melo, and L.~Montesano, ``Affordance-based imitation learning
  in robots,'' in \emph{Intelligent Robots and Systems, 2007. IROS 2007.
  IEEE/RSJ International Conference on}.\hskip 1em plus 0.5em minus 0.4em\relax
  IEEE, 2007, pp. 1015--1021.

\bibitem{andries2018affordance}
M.~Andries, R.~O. Chavez-Garcia, R.~Chatila, A.~Giusti, and L.~M. Gambardella,
  ``Affordance equivalences in robotics: a formalism,'' \emph{Frontiers in
  Neurorobotics}, vol.~12, p.~26, 2018.

\bibitem{shiraki2014modeling}
Y.~Shiraki, K.~Nagata, N.~Yamanobe, A.~Nakamura, K.~Harada, D.~Sato, and D.~N.
  Nenchev, ``Modeling of everyday objects for semantic grasp,'' in \emph{Robot
  and Human Interactive Communication, 2014 RO-MAN: The 23rd IEEE International
  Symposium on}.\hskip 1em plus 0.5em minus 0.4em\relax IEEE, 2014, pp.
  750--755.

\bibitem{bessler2018owl}
D.~Be{\ss}ler, M.~Pomarlan, and M.~Beetz, ``Owl-enabled assembly planning for
  robotic agents,'' in \emph{Proceedings of the 17th International Conference
  on Autonomous Agents and MultiAgent Systems}.\hskip 1em plus 0.5em minus
  0.4em\relax International Foundation for Autonomous Agents and Multiagent
  Systems, 2018, pp. 1684--1692.

\bibitem{myers2015affordance}
A.~Myers, C.~L. Teo, C.~Ferm{\"u}ller, and Y.~Aloimonos, ``Affordance detection
  of tool parts from geometric features.'' in \emph{ICRA}, 2015, pp.
  1374--1381.

\bibitem{abelha2017learning}
P.~Abelha~Ferreira and F.~Guerin, ``Learning how a tool affords by simulating
  3d models from the web,'' in \emph{Proceedings of IEEE International
  Conference on Intelligent Robots and Systems (IROS 2017)}.\hskip 1em plus
  0.5em minus 0.4em\relax IEEE Press, 2017.

\bibitem{rana2017towards}
M.~A. Rana, M.~Mukadam, S.~R. Ahmadzadeh, S.~Chernova, and B.~Boots, ``Towards
  robust skill generalization: Unifying learning from demonstration and motion
  planning,'' in \emph{Conference on Robot Learning}, 2017, pp. 109--118.

\bibitem{fitzgerald2014representing}
T.~Fitzgerald, A.~K. Goel, and A.~L. Thomaz, ``Representing skill
  demonstrations for adaptation and transfer,'' in \emph{AAAI Symposium on
  Knowledge, Skill, and Behavior Transfer in Autonomous Robots}, 2014.

\bibitem{gajewski2018adapting}
P.~Gajewski, P.~Ferreira, G.~Bartels, C.~Wang, F.~Guerin, B.~Indurkhya,
  M.~Beetz, and B.~Sniezynski, ``Adapting everyday manipulation skills to
  varied scenarios,'' \emph{arXiv preprint arXiv:1803.02743}, 2018.

\end{thebibliography}
	
\end{document}